\documentclass[runningheads]{llncs}
\usepackage{graphicx}

\usepackage{tcolorbox}
\usepackage{listings}
\usepackage{tcolorbox}
\newtcolorbox{dialogbox}{
  arc=4mm,
  colback=blue!3,
  colframe=black,
  rounded corners,
  boxrule=0.5pt,
  fonttitle=\bfseries,
  coltitle=black,

}
\usepackage{adjustbox}
\definecolor{darkbrown}{HTML}{8B4513}
\definecolor{navyblue}{HTML}{000080}
\definecolor{darkgreen}{HTML}{008000}
\definecolor{purple}{HTML}{800080}
\definecolor{red}{HTML}{FF0000}
\usepackage{booktabs}
\usepackage{multirow}
\usepackage[normalem]{ulem}
\useunder{\uline}{\ul}{}
\usepackage{amsmath,amssymb,amsfonts}

\usepackage{algorithmic}
\usepackage{algorithm}
\usepackage{graphicx}
\usepackage{textcomp}
\usepackage{xcolor}
\usepackage{enumitem}
\usepackage{colortbl}
\usepackage{tabularx}
\usepackage{hyperref}

\usepackage{orcidlink}
\usepackage{marvosym}

\usepackage{xcolor}

\usepackage{xcolor}

\begin{document}

\title{Evolve Cost-aware Acquisition Functions Using Large Language Models}
%
%
\author{Yiming Yao\inst{1,2,3}\textsuperscript{(\Letter)}\orcidlink{0009-0000-7069-6304} \and
Fei Liu\inst{2,3}\orcidlink{0000-0001-6719-0409} \and
Ji Cheng\inst{2,3}\orcidlink{0000-0002-1123-6030} \and
Qingfu Zhang\inst{2,3}\textsuperscript{(\Letter)}\orcidlink{0000-0003-0786-0671}}
\authorrunning{Y. Yao et al.}
%
\institute{City University of Hong Kong (Dongguan), Dongguan523000, China \and
Department of Computer Science, City University of Hong Kong, Hong Kong \and
The City University of Hong Kong Shenzhen Research Institute, Shenzhen, China\\
\email{\{yimingyao3-c,fliu36-c,J.Cheng\}@my.cityu.edu.hk, qingfu.zhang@cityu.edu.hk}
}
%
\maketitle              
\begin{abstract}
Many real-world optimization scenarios involve expensive evaluation with unknown and heterogeneous costs. Cost-aware Bayesian optimization stands out as a prominent solution in addressing these challenges. To approach the global optimum within a limited budget in a cost-efficient manner, the design of cost-aware acquisition functions (AFs) becomes a crucial step. However, traditional manual design paradigm typically requires extensive domain knowledge and involves a labor-intensive trial-and-error process. This paper introduces EvolCAF, a novel framework that integrates large language models (LLMs) with evolutionary computation (EC) to automatically design cost-aware AFs. Leveraging the crossover and mutation in the algorithmic space, EvolCAF offers a novel design paradigm, significantly reduces the reliance on domain expertise and model training. The designed cost-aware AF maximizes the utilization of available information from historical data, surrogate models and budget details. It introduces novel ideas not previously explored in the existing literature on acquisition function design, allowing for clear interpretations to provide insights into its behavior and decision-making process. In comparison to the well-known EIpu and EI-cool methods designed by human experts, our approach showcases remarkable efficiency and generalization across various tasks, including 12 synthetic problems and 3 real-world hyperparameter tuning test sets.

\keywords{Cost-aware Bayesian optimization  \and Acquisition functions \and Large language models \and Evolutionary computation.}
\end{abstract}
\section{Introduction}
Bayesian optimization (BO) is a powerful tool for solving expensive optimization problems and has found wide application in many real-world scenarios \cite{turner2021bayesian,garnett2010bayesian,negoescu2011knowledge,frazier2016bayesian}. It typically employs a surrogate model to approximate the expensive function and well-designed acquisition functions (AFs) to select potential solutions in a sample-efficient manner. Popular acquisition functions include probability of improvement (PI)~\cite{kushner1964new}, expected improvement (EI)~\cite{movckus1975bayesian}, upper confidence bound (UCB)~\cite{srinivas2009gaussian}, knowledge gradient (KG)~\cite{frazier2008knowledge}, etc.

Vanilla BO typically sets the number of evaluations as the budget constraint, implicitly assuming a uniform evaluation cost in the design space \cite{lee2020cost}. However, this is rarely the case in many real-world applications, leading to the concept of cost-aware BO. For example, in hyperparameter optimization (HPO) tasks for machine learning models, the costs with different hyperparameter configurations may even differ in the order of magnitudes \cite{lee2020cost}. Under the budget constraint of accumulated costs, approaching the global optimum is a challenge for traditional AFs due to their unawareness of heterogeneous evaluation costs, which highlights the importance of designing efficient cost-aware AFs.

Previous works have proposed several heuristics to take the cost information into account \cite{snoek2012practical,lee2020cost,lee2021nonmyopic}. Representative ones include EI per unit cost (EIpu) \cite{snoek2012practical} which divides EI by the cost function to promote solutions with both low cost and high improvement, and EI-cool \cite{lee2020cost} that introduces the cost-cooling strategy to make EIpu adapt to problems with expensive global optimum. Based on EIpu and EI-cool, several enhanced approaches have been suggested \cite{luong2021adaptive,qian2021cobabo,guinet2020pareto}. However, designing these AFs typically necessitates significant involvement of domain experts and extensive trial-and-error testing to refine and improve upon previous methods. This manual design paradigm is labor-intensive and non-automated. Furthermore, the information currently used to define cost-aware AFs is inadequate, as it only considers the EI metric and budget details while overlooking the complete historical data, which severely restricts the exploration in the algorithmic space. Simply integrating the EI metric with budget information does not inherently stimulate innovative ideas, thereby greatly limiting the performance and generalization of the designed AFs. While some model-based methods have been proposed to automatically learn AFs parameterized by neural networks in meta-BO community \cite{volpp2019meta,hsieh2021reinforced,maraval2024end}, they often require substantial effort in complex framework design and model training. Additionally, the resulting AFs are represented by network parameters, leading to poor interpretability compared to widely used AFs that have explicit mathematical expressions. Besides, these methods are designed for problems with uniform costs and are not applicable to many real-world applications with heterogeneous costs.

In the past three years, large language models (LLMs) have been widely used in code generation \cite{lehman2022evolution,liventsev2023fully,hemberg2024evolving}, mathematical reasoning \cite{romera2024mathematical} and automatic algorithm design \cite{liu2024evolution,liu2023algorithm}. While recent studies have explored the use of LLMs to enhance vanilla BO \cite{zhang2023using,liu2024large}, these approaches rely on querying LLMs to directly suggest candidate solutions. The absence of a clearly defined search strategy results in inadequate explainability. Moreover, whenever a new problem arises, it needs to conduct a substantial number of queries for LLMs from scratch, which can be expensive and impractical for real-world applications. 

We propose a novel paradigm, named EvolCAF, which integrates LLMs in an evolutionary framework to automatically design explicit AFs to enhance cost-aware BO. Different from the existing works, it enjoys good automation and explainability outperforming existing human-crafted methods. To the
best of our knowledge, this is the first attempt to utilize LLMs for automatic AF design for Bayesian optimization. Our main contributions are as follows:

\begin{itemize}
\item We introduce EvolCAF, which integrates large language models (LLMs) with evolutionary computation (EC) to automatically design cost-aware AFs. It enables crossover and mutation in the algorithmic space to iteratively search for elite AFs, significantly reducing the reliance on expert knowledge and domain model training.
\item We leverage EvolCAF to design a cost-aware AF that fully utilizes the available history information. Remarkably, the designed AF introduces novel ideas that have not been explored in existing literature on acquisition function design. The designed AF can be expressed explicitly, allowing for clear interpretations to provide insights into its behavior and decision-making process.
\item We evaluate the designed AF on diverse synthetic functions as well as practical hyperparameter optimization (HPO) problems. Compared to the popular EIpu and EI-cool methods designed by domain experts, our approach demonstrates remarkable efficiency and generalization, which highlights the promising potential in addressing many related real-world applications.
\end{itemize}

\section{Background and Related Works}
\subsection{Background}
In vanilla Bayesian optimization (BO), we consider finding the optimal solution $\mathbf{x}^*$ that maximizes the black-box objective function $f$: $\mathbf{x}^*=\arg\max_{\mathbf{x}\in\mathcal{X}}f(\mathbf{x})$, where $\mathcal{X}$ is a compact subset of $\mathbb{R}^d$, we assume $f:\mathcal{X}\to\mathbb{R}$ is continuously differentiable and expensive to evaluate.

\subsubsection{Gaussian Processes}
To approximate the expensive objective function, BO typically employs a Gaussian process (GP) model [1] as the surrogate. A GP is an infinite distribution over functions $f$ specified by a prior mean function $\mu(\cdot)$ and covariance function $k(\cdot, \cdot)$: $f(\mathbf{x}) \sim \mathcal{GP}_{f}\left(\mu(\mathbf{x}), k\left(\mathbf{x}, \mathbf{x}'\right)\right)$. Suppose in iteration $t$, the historical data set $\mathcal{D}_t=\{(\mathbf{x}_i,y_i)\}_{i=1}^t$ are obtained from the observation model $y_i=f(\mathbf{x}_i)+\epsilon_i$ with observation noise $\epsilon_i \sim \mathcal{N}\left(0, \sigma_{\epsilon}^{2}\right)$. Given the test point $\mathbf{x}^{*}$, the predictive distribution $p(y|\mathbf{x}^{*},\mathcal{D}_t)$ 
is also Gaussian with mean $\mu(\mathbf{x}^{*})=K_{\mathbf{x}^{*}, \mathbf{X}}(K_{\mathbf{X}, \mathbf{X}}+\sigma_{\epsilon}^{2} \mathrm{I})^{-1} \mathbf{y}$ and variance $\sigma^2(\mathbf{x}^{*})=k(\mathbf{x}^{*}, \mathbf{x}^{*})-K_{\mathbf{x}^{*}, \mathbf{X}}(K_{\mathbf{X}, \mathbf{X}}+\sigma_{\epsilon}^{2} \mathrm{I})^{-1} K_{\mathbf{X}, \mathbf{x}^{*}}$, where $\mathbf{y}=\left[y_{1}, y_{2}, \cdots, y_{t}\right]^{T}$ are noisy output values observed from the latent functions $\mathbf{f}=[f(\mathbf{x}_{1}), f(\mathbf{x}_{2}), \cdots, f(\mathbf{x}_{t})]^{T}$ at training points $\mathbf{X}=\left[\mathbf{x}_{1}, \mathbf{x}_{2}, \cdots, \mathbf{x}_{t}\right]^{T}$, $K_{\mathbf{X},\mathbf{X}}=\left[k\left(\mathbf{x}_{i}, \mathbf{x}_{j}\right)\right]_{\mathbf{x}_{i}, \mathbf{x}_{j} \in \mathbf{X}}$ is the covariance matrix and $K_{\mathbf{X},\mathbf{x}^{*}}=\left[k\left(\mathbf{x}_{i}, \mathbf{x}^{*}\right)\right]_{\mathbf{x}_{i} \in \mathbf{X}}$ is the correlation vector for all training and test points.\\
\subsubsection{Acquisition Functions} 
The acquisition function (AF) defines a utility that measures the benefit of evaluating an unknown point $\mathbf{x}$. We denote the definition of AF as $\alpha(\mathbf{x})$, which may contain the historical data set $\mathcal{D}_{t}$, model information $\mu(\mathbf{x})$, $\sigma^2(\mathbf{x})$, etc. One of the popular AFs is expected improvement (EI)~\cite{movckus1975bayesian}, which quantifies the expected amount of improvement over the current best observation $y^*=\max_iy_i$ at a given point $\mathbf{x}$ in the search space:
\begin{equation}
\alpha_{\mathrm{EI}}(\mathbf{x})=\mathbb{E}_{f(\mathbf{x})}[\left[f(\mathbf{x})-y^*]_+\right]=\sigma(\mathbf{x})\mathrm{~}h(\frac{\mu(\mathbf{x})-y^*}{\sigma(\mathbf{x})}),
\end{equation}
where $\mathbb{[}\cdot]_{+}$ is the $\max(0,\cdot)$ operation, $\mathrm{~}h(z)=\phi(z)+z\Phi(z)$, $\phi$ and $\Phi$ are the standard normal density and distribution functions, respectively. 

\subsection{Cost-aware Bayesian Optimization}
\subsubsection{Problem Setting} In cost-aware Bayesian optimization, it is assumed that evaluating the objective function is expensive. Additionally, the evaluation in different regions will incur heterogeneous costs, the unknown cost function is denoted as $c(\mathbf{x})$. For every query $\mathbf{x}_i$, we can obtain the noisy observation $y_i$ with cost $z_i=c(\mathbf{x}_i)+\eta_i$, where $\eta_i \sim \mathcal{N}\left(0, \sigma_{\eta}^{2}\right)$. Similar to the objective function $f$, the black-box cost function is modeled
as a draw from the Gaussian process $c(\mathbf{x})\sim\mathcal{GP}_{c}$. We use the posterior predictive mean of $\mathcal{GP}_{c}$ for calculating the cost function as \cite{snoek2012practical} and \cite{luong2021adaptive} do.

Given the historical data set $\bar{\mathcal{D}}_t=\{(\mathbf{x}_i,y_i,z_i)\}_{i=1}^t$ with $t$ evaluated samples and the limited total budget $B_{\mathrm{total}}$, we can only find a near-optimal solution with the constraint of cumulative cost $\sum_{i=1}^Tz_i\leq B_{\mathrm{total}}$, where $T$ is the maximum number of evaluated samples that satisfies the budget constraint. The general framework followed by the vast majority of existing cost-aware BO methods is shown in Algorithm \ref{cost-aware BO framework}, the main difference lies in the different definitions of cost-aware AFs, which will be introduced below.

\begin{algorithm}[H]
	\renewcommand{\algorithmicrequire}{\textbf{Input:}}
	\renewcommand{\algorithmicensure}{\textbf{Output:}}
	\caption{Cost-aware BO}
	\label{cost-aware BO framework}
	\begin{algorithmic}[1]
            \REQUIRE~\\
            $B_{\mathrm{total}}$: total budget,  $\bar{\mathcal{D}}_{t}$: initial data set with $t$ evaluated samples
            \STATE Initialize used budget $B_{\mathrm{used}} =\sum_{i=1}^tz_i$ \\
            \STATE Train objective and cost models $\mathcal{GP}_{f}$ and $\mathcal{GP}_{c}$ using $\bar{\mathcal{D}}_{t}$ \\ 
            \WHILE{$B_{\mathrm{used}}<B_{\mathrm{total}}$}
            \STATE Query candidate: $\mathbf{x}_{t+1}=\arg\max_\mathbf{x}\alpha(\mathbf{x})$ \\
            \STATE Evaluate candidate:
            $y_{t+1},z_{t+1}\leftarrow f(\mathbf{x}_{t+1}),c(\mathbf{x}_{t+1})$ \\
            \STATE Update data set 
            $\bar{\mathcal{D}}_{t+1}\leftarrow\bar{\mathcal{D}}_{t}\cup\{(\mathbf{x}_{t+1},y_{t+1},z_{t+1})\}$ \\
            \STATE Update models $\mathcal{GP}_{f}$ and $\mathcal{GP}_{c}$ using $\bar{\mathcal{D}}_{t+1}$ \\
            \STATE $B_{\mathrm{used}}\leftarrow B_{\mathrm{used}}+z_{t+1}$ \\
            \STATE $t\leftarrow t+1$ \\
            \ENDWHILE
		\STATE Total number of evaluated samples $T\leftarrow t$ \\
		\ENSURE Best configuration $\arg\max_{(\mathbf{x}_i,y_i)\in \bar{\mathcal{D}}_{T}}y_i$
	\end{algorithmic}  
\end{algorithm}

\subsubsection{EI per Unit Cost (EIpu)}
To balance the cost and quality of evaluations, Snoek et al.~\cite{snoek2012practical} proposed EI per unit cost (EIpu), which normalizes the EI metric by the cost function $c(\mathbf{x})$:
\begin{equation}
\alpha_\mathrm{EIpu}(\mathbf{x})=\frac{\alpha_\mathrm{EI}(\mathbf{x})}{c(\mathbf{x})}.
\label{eq:EIpu}
\end{equation}

By using the cost function to penalize EI, EIpu tends to carefully select candidate points with low cost and high improvement, making the search process cost-aware. Benefiting from the preference for cheaper regions, EIpu can be consistently improved when the optimum is cheap to evaluate since higher EI and lower cost are both encouraged. However, the preference becomes a drawback when the optimum lies in the expensive regions of the design space, which is common in many real-world applications. The cost penalty term prevents EIpu from exploring near-optimal regions that are expensive to evaluate, experiments have shown that sometimes EIpu performs even worse than EI \cite{lee2020cost}. 

\subsubsection{EI-cool}
To alleviate the above problem, Lee et al.~\cite{lee2020cost} introduced a cost-cooling factor $\alpha$ in EIpu called EI-cool:
\begin{equation}
\alpha_\mathrm{EI-cool}(\mathbf{x})=\frac{\alpha_\mathrm{EI}(\mathbf{x})}{c(\mathbf{x})^\alpha},
\label{eq:EI-cool}
\end{equation}
where $\alpha=(B_{\mathrm{total}}-B_{\mathrm{used}})/(B_{\mathrm{total}}-B_\mathrm{init})$, $B_\mathrm{init}$ is the budget spent in evaluating the initial sample points, $B_{\mathrm{used}}$ is the budget already used and $B_{\mathrm{total}}$ is the given total budget. As $B_{\mathrm{used}}$ increases from $B_\mathrm{init}$ to $B_{\mathrm{total}}$ during the search process, the factor $\alpha$ gradually decays from 1 to 0, resulting in the transition of EI-cool from EIpu to EI. Intuitively, the cost-cooling strategy diminishes the significance of the cost model as the budget is consumed, making EI-cool to operate in an \textit{early and cheap, late and expensive} fashion to encourage exploring expensive regions when the remaining budget is tight.

Although EI-cool alleviates the problem of performance degradation when searching for the expensive optimum, it always uses EIpu as the starting strategy, which is not flexible and may not adapt well to different problems~\cite{luong2021adaptive}. Besides, previous analysis and experiments have shown that when the remaining budget is gradually tight, although deemphasizing the cost can increase the likelihood of exploring expensive regions, it is still possible to miss the optimum in high-cost regions before the budget is exhausted \cite{luong2021adaptive}, as the exploration or exploitation in very cheap regions can still result in very large values of EI-cool metric, which is called low-cost-preference weakness in \cite{qian2021cobabo}.

\subsubsection{Variants Based on EIpu and EI-cool}
Based on EIpu and EI-cool, some improved methods have been proposed such as using a multi-armed bandit algorithm to automatically select either EI or EIpu \cite{luong2021adaptive}, developing more aggressive methods to alleviate the low-cost-preference weakness of EI-cool\cite{qian2021cobabo}, and cost-aware EI based on Pareto optimality to achieve the trade-off between cost and improvement \cite{guinet2020pareto}. It is evident that the design process typically requires significant involvement of domain knowledge and extensive trial-and-error testing based on the flaws of previous methods, which are labor-intensive and non-automated. Besides, the existing AFs just combine the EI metric with budget information in different ways, which severely restricts the exploration in the algorithmic space and does not inherently foster the generation of innovative ideas, so the performance and generalization are greatly limited.

\subsection{Automatic Design for Acquisition Functions}
In the meta-BO community, which focuses on meta-learning or learning to learn to enhance vanilla BO \cite{chen2017learning,tv2019meta,chen2022towards,bai2023transfer}, there has been research dedicated to automatically generating efficient and generalizable AFs via a learning model. While these works are not tailored for cost-aware contexts, we will review them to emphasize the strengths and potential of our framework.

To learn a meta-acquisition function, Volpp et al. \cite{volpp2019meta} replaced the hand-designed AF with a neural network named neural acquisition function (NAF), which is meta-trained on related source tasks by policy-based reinforcement learning. Hsieh et al. \cite{hsieh2021reinforced} utilized a deep Q-network (DQN) as a surrogate differentiable AF to achieve a few-shot fast adaptation of AFs. Maraval et al. \cite{maraval2024end} introduced an end-to-end differentiable framework based on transformer architectures called neural acquisition process (NAP) to meta-learn acquisition functions with the surrogate model jointly. Nevertheless, despite achieving promising results, these model-based methods often demand substantial effort in framework design and model training. Moreover, the resulting AFs are represented by network parameters, resulting in poor interpretability compared to widely used AFs that have explicit mathematical expressions.

\section{EvolCAF: Evolve Cost-aware Acquisition Functions with LLMs}
\subsection{Framework}
The proposed EvolCAF framework embraces the basic components of evolutionary computing (EC), including initialization, crossover, mutation, and population management. In EvolCAF, each individual represents an acquisition function solving a branch of synthetic instances, which is represented with an algorithm description and a code block implementation instead of an encoded vector in traditional EC. During the evolution process, the initialization, crossover, and mutation operations on the individuals are all performed by prompting LLM in the algorithmic space. The entire process is completely automated without any intervention from human experts.

\begin{figure}
\centering
\includegraphics[scale=0.3]{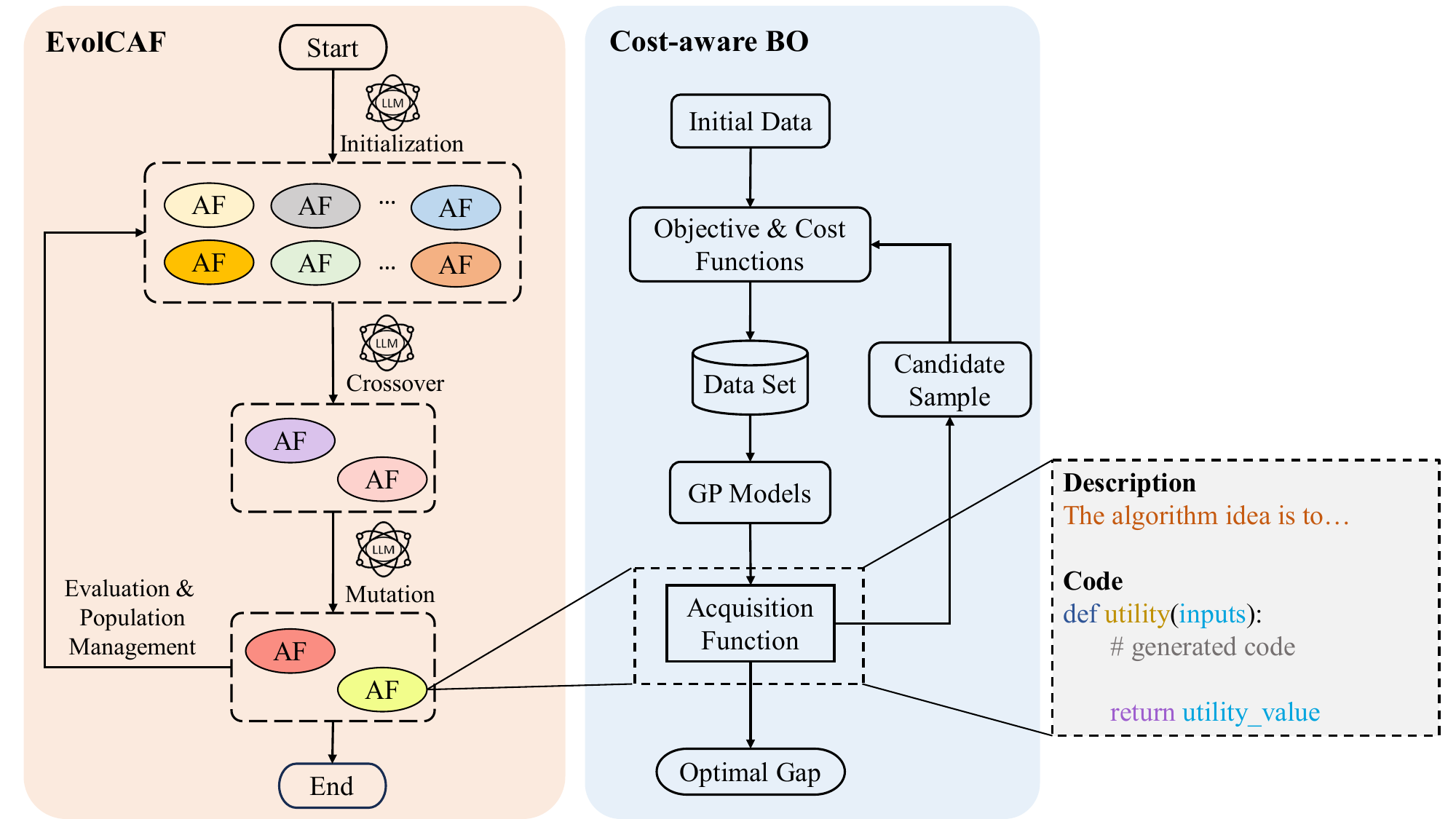}
\caption{Flowchart of EvolCAF framework. The left box presents the evolution of cost-aware AFs enabled by EvolCAF, wherein each individual in the population is an AF, represented with an algorithm description and a code block implementation. The initialization, crossover, and mutation are facilitated by LLMs. The middle box shows the cost-aware BO loop, each AF is evaluated on a set of synthetic instances to calculate the optimal gap as its fitness value.} \label{EvolCAF}
\end{figure}

Fig.~\ref{EvolCAF} illustrates the detailed flowchart of EvolCAF. At each generation, we maintain a population of $N$ AFs, each AF is evaluated on a set of synthetic instances in a cost-aware BO loop to calculate the fitness value, which is the optimal gap between the true optimal value and the optimal value obtained by the AF. After new individuals are added to the population, the worst individuals are deleted according to the fitness values. 

\subsection{General Definition for Evolved AFs}
The general definition for evolved cost-aware AFs can be formulated as follows:
\begin{equation}
\begin{split}
&\alpha_{\mathrm{EvolCAF}}(\mathbf{x})=\alpha(\mathbf{x};\boldsymbol{\theta}_\mathrm{data},\boldsymbol{\theta}_\mathrm{model},\boldsymbol{\theta}_\mathrm{budget}) \\
&=\alpha(\mathbf{x};
\mathbf{X},\mathbf{y},\mathbf{x}^{*},{y}^{*},
\mu(\mathbf{x}),\sigma(\mathbf{x}),
c(\mathbf{x}),B_{\mathrm{used}},B_{\mathrm{total}}).
\end{split}
\label{eq:architecture for EvolCAF}
\end{equation}

The inputs of cost-aware AFs incorporate three groups of information: (1) historical data $\boldsymbol{\theta}_\mathrm{data}$=$\{\mathbf{X},\mathbf{y},\mathbf{x}^{*},{y}^{*}\}$, 
(2) prediction and uncertainty provided by the model $\boldsymbol{\theta}_\mathrm{model}$=$\{\mu(\mathbf{x}),\sigma(\mathbf{x})\}$, and (3) budget information during the optimization  $\boldsymbol{\theta}_\mathrm{budget}$=$\{c(\mathbf{x}),B_{\mathrm{used}},B_{\mathrm{total}}\}$. The first two groups include the data and model information for searching with uncertainties, while the last group informs the AF of the budget constraints in optimization. During the evolutionary process, EvolCAF is encouraged to explore the algorithmic space to generate and refine elite cost-aware AFs. 

\subsection{Prompt Engineering}
\begin{figure}[tbp]
\centering
\includegraphics[width=\textwidth]{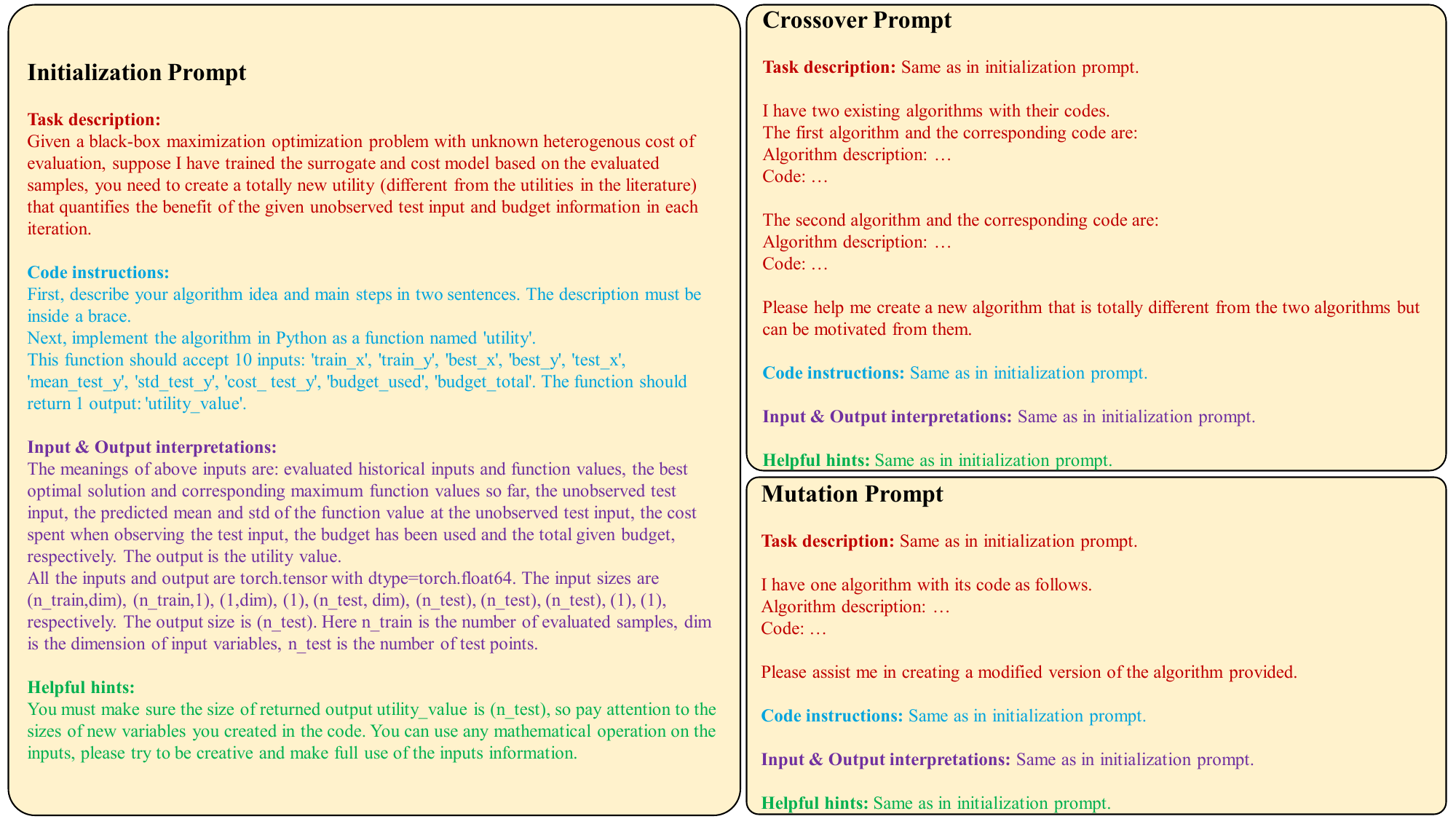}
\caption{Prompts used in EvolCAF for initialization, crossover, and mutation.} \label{prompt}
\end{figure}

The general format of prompt engineering used to inform LLMs consists of four parts: (1) a general description of the task, (2) code instructions for implementing algorithms, including the function name, inputs and output, (3) interpretations for the inputs and output, including their detailed meanings in our task, the variable formats and dimensions implemented in the code, (4) helpful hints to inform LLMs to generate executable codes and utilize input information as much as possible to create novel ideas. Following the general format, in initialization, we instruct LLMs to create a completely new AF to promote population diversity. In crossover, we suggest combining the selected parent AFs to facilitate the preservation of high-performing components in the following generations. While in mutation, we aim to encourage the exploration of better AFs based on parent AFs in the algorithmic space. The details of prompts for initialization, crossover, and mutation are shown in Fig.~\ref{prompt}.

\section{Experimental Studies}
\subsection{Experimental Settings}
\subsubsection{Settings for AF Evolution}
In the evolutionary process, EvolCAF maintains 10 AFs and evolves over 20 generations. We generate 1 offspring individual in each generation based on 2 parent individuals, with the crossover probability set to 1.0 and mutation probability set to 0.5. The GPT-3.5-turbo pre-trained LLM is used for generating AFs.

To generate AFs that can be efficiently optimized, we set a time threshold of 60 seconds for completing the cost-aware BO loop, which serves as a selection pressure together with the fitness value. Any AF that exceeds this time limit will be automatically eliminated during the evolutionary process.

\subsubsection{Settings for Cost-aware BO}
To calculate the fitness value, we evaluate each evolved AF on 2D Ackley and 2D Rastrigin functions with 10 different random seeds in the experimental design, resulting in a total of 20 instances. The aim is to achieve improved generalization results across various initial surrogate landscapes with an acceptable evaluation time during evolution, as the training and inference of GP models on a large number of instances in each generation can be expensive. The fitness value for each evolved AF is calculated by averaging the optimal gaps obtained from the 20 instances.

To simulate the scenarios in many real-world applications, we carefully design a cost function that is most expensive to evaluate at the global optimum $\mathbf{x}^*$ of the synthetic function, the formulation is similar to that used in \cite{lee2021nonmyopic} and can be expressed as:
\begin{equation}
c(\mathbf{x}) = \exp\left(-\left\| \mathbf{x} - \mathbf{x}^* \right\|_2\right),
\label{eq:cost function}
\end{equation}
where each dimension of $\mathbf{x}$ and $\mathbf{x}^*$ is normalized to [0,1]. In order to achieve good results for the evolved AF given a small budget, we set the total budget $B_{\mathrm{total}}$ as 30 in the evolutionary process, indicating that the smallest number of evaluations is 30, we will further verify the generalization using sufficient budget in the following experiments. We initialize $2d$ random samples using experimental design, where $d$ is the dimension of the decision variable.

All BO methods are implemented using BoTorch \cite{balandat2020botorch}. In the BO loop, the acquisition functions are optimized through multi-start optimization using scipy's L-BFGS-B optimizer, using 20 restarts seeded from 100 pseudo-random samples through BoTorch’s initialization heuristic for efficient optimization. 

\subsection{Evolution Results}
\begin{figure}[]
\centering
\includegraphics[scale=0.3]{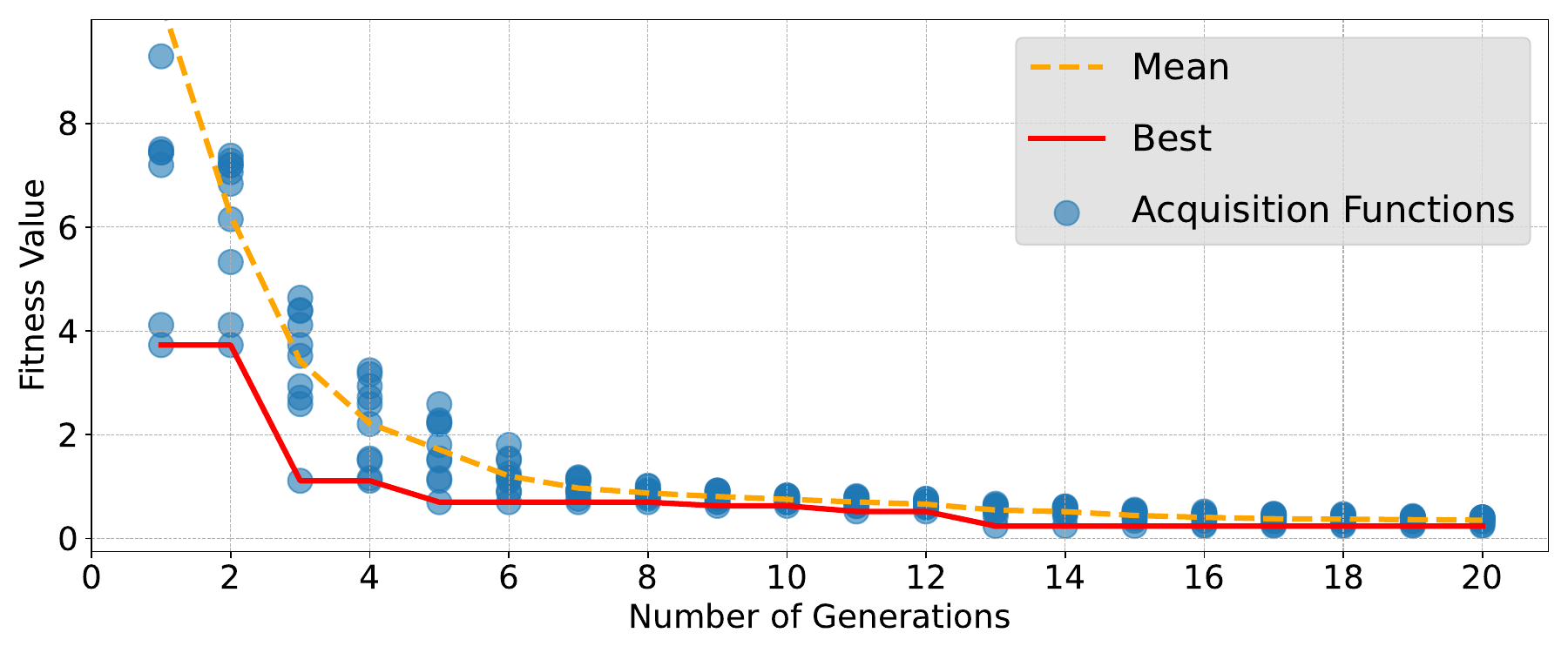}
\caption{The evolutionary process of acquisition functions.} 
\label{evolution process}
\end{figure}

Fig.~\ref{evolution process} demonstrates the evolutionary process. In each generation, we maintain 10 AFs represented by blue dots. The mean and optimal fitness values of the population are represented with orange and red lines, respectively. With a population size of 10 and 20 generations, the fitness value of the evolving AFs can converge to a notably low level. The results show the capability of our framework to automatically generate and evolve elite AFs.

\begin{figure}[thbp]
\centering
\includegraphics[width=\textwidth]{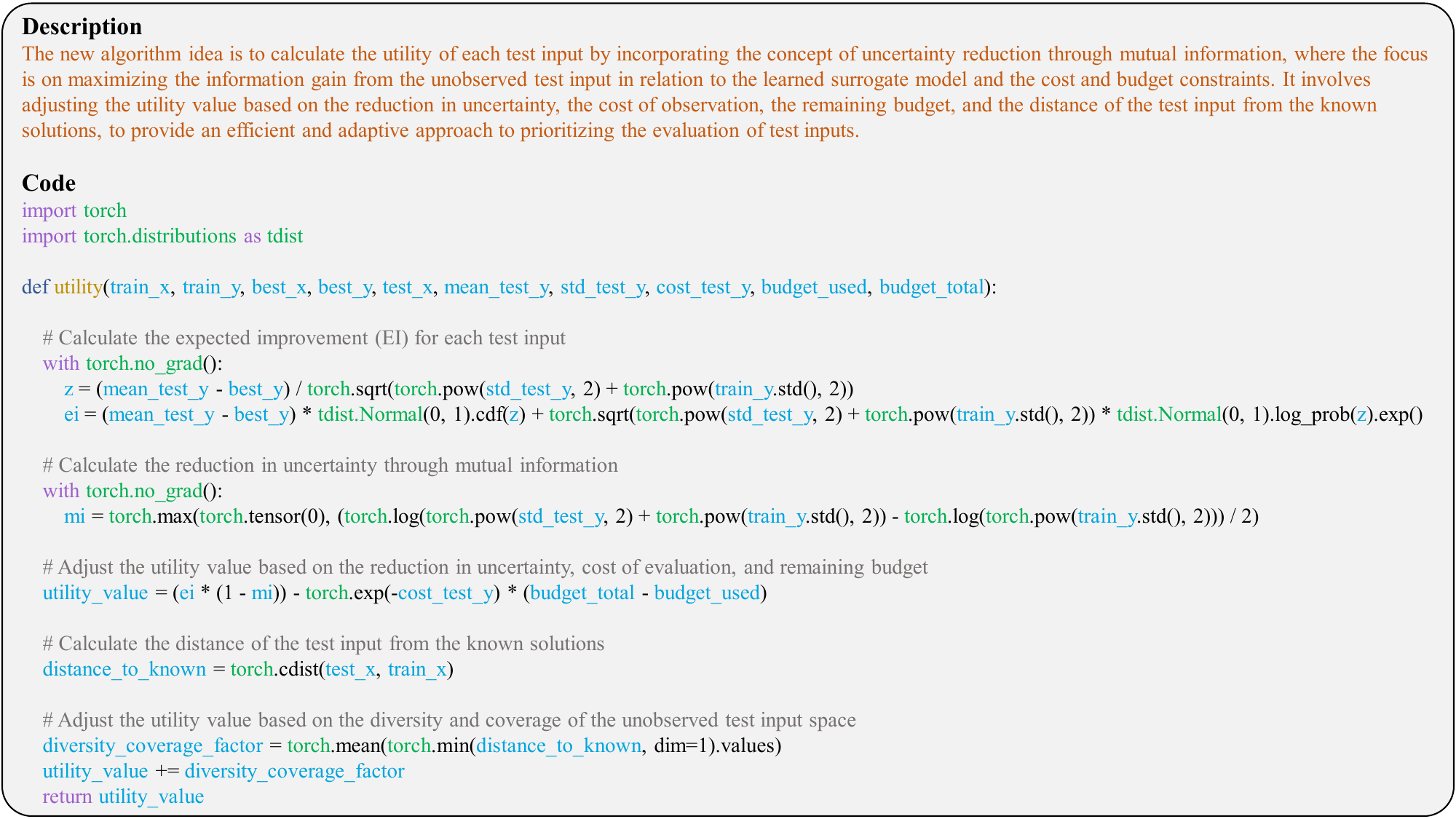}
\caption{Optimal acquisition function designed by EvolCAF. The results include a linguistic description of the algorithmic idea, as well as a code implementation with annotations, all the contents are produced by LLMs.} \label{optimal code}
\end{figure}

Fig.~\ref{optimal code} shows the optimal AF designed by EvolCAF with the minimum fitness value, including a general description of the algorithmic idea in defining the AF and a detailed code implementation. After converting the code into easily understandable mathematical expressions, we observe that the optimal AF consists of three parts:
\begin{equation}
\alpha_{\mathrm{EvolCAF}}(\mathbf{x})=\alpha_1(\mathbf{x})+\alpha_2(\mathbf{x})+\alpha_3(\mathbf{x}).
\label{eq:alpha}
\end{equation}

Specifically, $\alpha_1(\mathbf{x})$ combines a modified EI with uncertainty information:
\begin{equation}
\begin{split}
\alpha_1(\mathbf{x})=&[(\mu(\mathbf{x})-y^{*})\Phi(\mathbf{z})+\sqrt{\sigma^2(\mathbf{x})+\sigma^2(\mathbf{y})}\cdot\phi(\mathbf{z})]\cdot\\&(1-\log\sqrt{\frac{\sigma^2(\mathbf{x})+\sigma^2(\mathbf{y})}{\sigma^2(\mathbf{y})}}),
\end{split}
\label{eq:alpha 1}
\end{equation}
where $\mathbf{z}=\frac{\mu(\mathbf{x})-y^{*}}{\sqrt{\sigma^2(\mathbf{x})+\sigma^2(\mathbf{y})}}$, $\phi$ and $\Phi$ are the standard normal density and distribution functions, respectively, $\sigma^2(\mathbf{y})$ represents the variance of the current historical observations. 

Similar to EI, $\alpha_1(\mathbf{x})$ encourages searching regions close to the current best observation with uncertainties. However, an improvement is that $\alpha_1(\mathbf{x})$ also incorporates the uncertainty of all historical observations rather than solely focusing on the current best observation value and the uncertainty of the unknown point $\mathbf{x}$.

$\alpha_2(\mathbf{x})$ mainly focuses on the current remaining budget and the cost of evaluating the unknown point $\mathbf{x}$:
\begin{equation}
\alpha_2(\mathbf{x})=-\frac{B_{\mathrm{total}}-B_{\mathrm{used}}}{e^{c(\mathbf{x})}}.
\label{eq:alpha 2}
\end{equation}

It can be observed that $\alpha_2(\mathbf{x})$ enables the optimization to focus on EI regardless of the cost when the remaining budget is tight, which is similar to the cost-cooling strategy used in EI-cool. However, the difference is that $\alpha_2(\mathbf{x})$ will not only keep the optimization encouraging higher EI metrics but also promote the exploration of expensive regions when there is a sufficient budget. This feature addresses the low-cost-preference weakness in EI-cool, allowing for a more comprehensive search.

$\alpha_{3}(\mathbf{x})$ considers the distance between the unknown inputs and historical observed locations when optimizing AF:
\begin{equation}
\alpha_{3}(\mathbf{x})=\frac{1}{m}\sum_{i=1}^{m}\min_{j} A_{ij},
\label{eq:alpha 3}
\end{equation}
where $A_{ij}=dist(\mathbf{u}_i,\mathbf{v}_j)$ is the distance matrix, $\{\mathbf{u}_i\}_{i=1}^m$ are multiple starting points used in the efficient multi-start optimization scheme of BoTorch, $\mathbf{v}_{j}$ is the $j$th element in observed locations $\mathbf{X}$. Therefore, $\alpha_{3}(\mathbf{x})$ utilizes the information of all the distances between the unobserved location in each optimization trajectory and observed ones in the historical data set. When optimizing $\alpha_{\mathrm{EvolCAF}}(\mathbf{x})$, maximizing the average of the minimum distances contributed by $\alpha_{3}(\mathbf{x})$ forces the multi-start optimization away from explored regions.

The analyses above suggest that the designed AF can introduce novel ideas that have not been previously explored in existing literature on acquisition function design. Benefiting from the evolution in the algorithmic space, the designed AF can be expressed explicitly, allowing for clear interpretations to provide insights into its behavior and decision-making process.

\subsection{Evaluation of the Optimal Acquisition Function}
\subsubsection{Synthetic Problems}
In this subsection, we evaluate the optimal AF on 12 different synthetic instances with different landscapes and input dimensions. We define the cost functions according to Equation (\ref{eq:cost function}). As the evolution is conducted within a total budget of 30, which is to enable the optimal AF to achieve better results using a small budget, therefore, we also tested the generalization of the optimal AF within a sufficient budget of 300. Each test instance is conducted with 10 independent runs, the results are shown in Table \ref{synthetic exp}.

\begin{table}
\centering
\caption{Means of optimal gaps (number of evaluated samples) obtained by different AFs on all synthetic instances over 10 independent runs. The best mean result for each row is highlighted in bold. }\label{synthetic exp}
\scalebox{0.75}{
\begin{tabular}{cccccc}
\toprule[1pt]
Test Instances                     & Budget & EI             & EIpu                    & EI-cool                 & EvolCAF              \\ \hline
\multirow{2}{*}{Ackley 2D}         & 30   & 2.6600(40)     & 2.3302(40)              & 2.7369(40)              & \textbf{0.4277(34)}  \\
                                   & 300  & 1.2295(395)    & 0.8582(399)             & 0.8317(399)             & \textbf{0.0505(306)} \\ \hline
\multirow{2}{*}{Rastrigin 2D}      & 30   & 4.7425(41)     & 5.6155(41)              & 5.7754(40)              & \textbf{0.0511(34)}  \\
                                   & 300  & 1.6656(410)    & 1.6678(408)             & 1.8518(408)             & \textbf{0.0046(306)} \\ \hline
\multirow{2}{*}{Griewank 2D}       & 30   & 0.4875(35)     & 0.3384(36)              & 0.3374(36)              & \textbf{0.1762(33)}  \\
                                   & 300  & 0.1305(323)    & 0.1195(323)             & 0.1360(323)             & \textbf{0.0361(307)} \\ \hline
\multirow{2}{*}{Rosenbrock 2D}     & 30   & 1.2609(41)     & 2.3601(44)              & 2.2909(42)              & \textbf{0.0304(33)}  \\
                                   & 300  & 0.0332(369)    & 0.0406(394)             & \textbf{0.0317(372)}    & 0.0402(307)          \\ \hline
\multirow{2}{*}{Levy 2D}           & 30   & 0.0056(38)     & 0.0098(38)              & 0.0116(38)              & \textbf{0.0013(33)}  \\
                                   & 300  & 1.1517e-4(314) & \textbf{5.9321e-5(316)} & 8.1046e-5(317)          & 3.7248e-4(307)       \\ \hline
\multirow{2}{*}{ThreeHumpCamel 2D} & 30   & 0.0483(39)     & 0.1182(40)              & 0.0710(39)              & \textbf{0.0007(33)}  \\
                                   & 300  & 5.0446e-4(322) & 7.4557e-4(326)          & \textbf{2.6392e-4(325)} & 7.5310e-4(306)       \\ \hline
\multirow{2}{*}{StyblinskiTang 2D} & 30   & 0.0286(41)     & 0.0233(42)              & 0.0266(41)              & \textbf{0.0071(33)}  \\
                                   & 300  & 1.4420e-4(332) & 1.8616e-4(339)          & \textbf{6.1798e-5(343)} & 2.0142e-3(306)       \\ \hline
\multirow{2}{*}{Hartmann 3D}       & 30   & 5.6696e-5(40)  & 1.0364e-4(41)           & \textbf{4.6158e-5(40)}  & 4.8127e-4(36)        \\
                                   & 300  & 1.8263e-5(420) & 1.3089e-5(429)          & \textbf{9.0599e-6(432)} & 2.3656e-4(311)       \\ \hline
\multirow{2}{*}{Powell 4D}         & 30   & 18.8892(48)    & 19.8281(51)             & 14.9481(49)             & \textbf{0.1285(38)}  \\
                                   & 300  & 2.9839(376)    & 1.1173(395)             & 1.6806(391)             & \textbf{0.0136(316)} \\ \hline
\multirow{2}{*}{Shekel 4D}         & 30   & 7.9123(48)     & 7.9210(49)              & 8.2132(48)              & \textbf{2.6367(39)}  \\
                                   & 300  & 6.5193(545)    & 6.9044(545)             & 7.0135(551)             & \textbf{0.1993(315)} \\ \hline
\multirow{2}{*}{Hartmann 6D}       & 30   & 0.0326(52)     & 0.0296(52)              & \textbf{0.0278(52)}     & 0.0384(44)           \\
                                   & 300  & 0.0122(710)    & 0.0054(705)             & 0.0154(695)             & \textbf{0.0042(327)} \\ \hline
\multirow{2}{*}{Cosine8 8D}        & 30   & 0.4723(48)     & 0.4738(48)              & 0.5351(48)              & \textbf{0.4357(53)}  \\
                                   & 300  & 0.1707(532)    & 0.2364(533)             & 0.2779(527)             & \textbf{0.0148(342)} \\ 
\bottomrule[1pt]
\end{tabular}}
\vspace{-1.8em} 
\end{table}

Based on the experimental results, it can be observed that in the vast majority of cases, the optimal AF achieves significantly better performance than EI and other cost-aware AFs. The optimal AF demonstrates strong generalization capabilities across unseen instances with diverse landscapes and sufficient budget constraints. In addition to the promising performance, an interesting phenomenon is that within a fixed budget, the optimal AF uses fewer evaluations in most cases, which is more pronounced when the budget is sufficient. 

To verify the contribution of each component in the optimal AF, we further display the performance of EvolCAF after removing each of the three components, as shown in Table \ref{synthetic ablation}. It can be observed that removing $\alpha_2(\mathbf{x})$, which takes into account budget information, has the greatest impact on the final performance of EvolCAF. Compared with EI, EIpu and EI-cool, the cost-aware AFs that remove $\alpha_1(\mathbf{x})$ or $\alpha_3(\mathbf{x})$ can still achieve better results. The results indicate the effectiveness and superiority of the designed acquisition function. 

\begin{table}
\centering
\caption{Means of optimal gaps (number of evaluated samples) obtained by different AFs on all synthetic instances over 10 independent runs. The best, second best, and worst mean results for each row are highlighted with bold fonts, underlines, and shaded background, respectively.}\label{synthetic ablation}
\scalebox{0.63}{
\begin{tabular}{ccccccccc}
\toprule[1pt]
Test Instances                      & Budget & EI                                  & EIpu                               & EI-cool                             & w/o alpha1           & w/o alpha2                          & w/o alpha3                          & EvolCAF               \\ \hline
                                    & 30   & 2.6600(40)                          & 2.3302(40)                         & 2.7369(40)                          & {\ul 0.6422(34)}     & \cellcolor[HTML]{C0C0C0}4.8046(40)  & 1.5008(33)                          & \textbf{0.4277(34)}  \\ 
\multirow{-2}{*}{Ackley 2D}         & 300  & 1.2295(395)                         & 0.8582(399)                        & 0.8317(399)                         & 0.7301(308)          & \cellcolor[HTML]{C0C0C0}1.2677(345) & \textbf{0.0501(305)}                & {\ul 0.0505(306)}    \\ \hline
                                    & 30   & 4.7425(41)                          & 5.6155(41)                         & 5.7754(40)                          & {\ul 0.0746(34)}     & \cellcolor[HTML]{C0C0C0}5.7126(42)  & 0.2648(33)                          & \textbf{0.0511(34)}  \\ 
\multirow{-2}{*}{Rastrigin 2D}      & 300  & 1.6656(410)                         & 1.6678(408)                        & \cellcolor[HTML]{C0C0C0}1.8518(408) & 0.0842(308)          & 0.8550(383)                         & {\ul 0.0157(306)}                   & \textbf{0.0046(306)} \\ \hline
                                    & 30   & 0.4875(35)                          & 0.3384(36)                         & 0.3374(36)                          & {\ul 0.1913(34)}     & \cellcolor[HTML]{C0C0C0}0.6671(36)  & 0.3535(33)                          & \textbf{0.1762(33)}  \\ 
\multirow{-2}{*}{Griewank 2D}       & 300  & 0.1305(323)                         & 0.1195(323)                        & 0.1360(323)                         & {\ul 0.0522(308)}    & \cellcolor[HTML]{C0C0C0}0.1954(324) & 0.0598(306)                         & \textbf{0.0361(307)} \\ \hline
                                    & 30   & 1.2609(41)                          & \cellcolor[HTML]{C0C0C0}2.3601(44) & 2.2909(42)                          & {\ul 0.0399(33)}     & 2.1626(39)                          & 2.3467(33)                          & \textbf{0.0304(33)}  \\ 
\multirow{-2}{*}{Rosenbrock 2D}     & 300  & 0.0332(369)                         & 0.0406(394)                        & {\ul 0.0317(372)}                   & \textbf{0.0104(308)} & 0.0587(348)                         & \cellcolor[HTML]{C0C0C0}1.8554(307) & 0.0402(307)          \\ \hline
                                    & 30   & 0.0056(38)                          & 0.0098(38)                         & 0.0116(38)                          & \textbf{0.0006(34)}  & \cellcolor[HTML]{C0C0C0}0.0335(37)  & 0.0195(33)                          & {\ul 0.0013(33)}     \\ 
\multirow{-2}{*}{Levy 2D}           & 300  & 1.1517e-4(314)                      & \textbf{5.9321e-5(316)}            & {\ul 8.1046e-5(317)}                & 0.0003(307)          & 0.0009(327)                         & \cellcolor[HTML]{C0C0C0}0.0010(306) & 3.7248e-4(307)       \\ \hline
                                    & 30   & 0.0483(39)                          & \cellcolor[HTML]{C0C0C0}0.1182(40) & 0.0710(39)                          & \textbf{0.0006(34)}  & 0.0940(38)                          & 0.0188(33)                          & {\ul 0.0007(33)}     \\ 
\multirow{-2}{*}{ThreeHumpCamel 2D} & 300  & {\ul 5.0446e-4(322)}                & 7.4557e-4(326)                     & \textbf{2.6392e-4(325)}             & 0.0005(308)          & 0.0020(332)                         & \cellcolor[HTML]{C0C0C0}0.0099(307) & 7.5310e-4(306)       \\ \hline
                                    & 30   & 0.0286(41)                          & 0.0233(42)                         & 0.0266(41)                          & {\ul 0.0136(33)}     & \cellcolor[HTML]{C0C0C0}1.7123(41)  & 0.0713(33)                          & \textbf{0.0071(33)}  \\ 
\multirow{-2}{*}{StyblinskiTang 2D} & 300  & {\ul 1.4420e-4(332)}                & 1.8616e-4(339)                     & \textbf{6.1798e-5(343)}             & 0.0069(307)          & \cellcolor[HTML]{C0C0C0}0.0246(332) & 0.0042(306)                         & 2.0142e-3(306)       \\ \hline
                                    & 30   & {\ul 5.6696e-5(40)}                 & 1.0364e-4(41)                      & \textbf{4.6158e-5(40)}              & 0.0007(36)           & \cellcolor[HTML]{C0C0C0}0.1438(51)  & 0.0731(36)                          & 4.8127e-4(36)        \\ 
\multirow{-2}{*}{Hartmann 3D}       & 300  & 1.8263e-5(420)                      & {\ul 1.3089e-5(429)}               & \textbf{9.0599e-6(432)}             & 0.0004(311)          & \cellcolor[HTML]{C0C0C0}0.0124(446) & 0.0005(311)                         & 2.3656e-4(311)       \\ \hline
                                    & 30   & 18.8892(48)                         & 19.8281(51)                        & 14.9481(49)                         & {\ul 0.1719(39)}     & \cellcolor[HTML]{C0C0C0}36.0514(56) & 4.3751(37)                          & \textbf{0.1285(38)}  \\ 
\multirow{-2}{*}{Powell 4D}         & 300  & \cellcolor[HTML]{C0C0C0}2.9839(376) & 1.1173(395)                        & 1.6806(391)                         & {\ul 0.0205(316)}    & 1.7473(450)                         & 0.2997(317)                         & \textbf{0.0136(316)} \\ \hline
                                    & 30   & 7.9123(48)                          & 7.9210(49)                         & 8.2132(48)                          & \textbf{2.3629(39)}  & \cellcolor[HTML]{C0C0C0}9.2281(60)  & 3.5714(37)                          & {\ul 2.6367(39)}     \\ 
\multirow{-2}{*}{Shekel 4D}         & 300  & 6.5193(545)                         & 6.9044(545)                        & 7.0135(551)                         & 0.3430(316)          & \cellcolor[HTML]{C0C0C0}8.1076(554) & \textbf{0.1583(313)}                & {\ul 0.1993(315)}    \\ \hline
                                    & 30   & 0.0326(52)                          & {\ul 0.0296(52)}                   & \textbf{0.0278(52)}                 & 0.0343(44)           & \cellcolor[HTML]{C0C0C0}1.7641(83)  & 0.1305(42)                          & 0.0384(44)           \\ 
\multirow{-2}{*}{Hartmann 6D}       & 300  & 0.0122(710)                         & 0.0054(705)                        & 0.0154(695)                         & {\ul 0.0044(326)}    & \cellcolor[HTML]{C0C0C0}0.4572(750) & 0.0127(327)                         & \textbf{0.0042(327)} \\ \hline
                                    & 30   & 0.4723(48)                          & 0.4738(48)                         & 0.5351(48)                          & {\ul 0.4438(53)}     & \cellcolor[HTML]{C0C0C0}1.5106(88)  & 0.6058(46)                          & \textbf{0.4357(53)}  \\ 
\multirow{-2}{*}{Cosine8 8D}        & 300  & 0.1707(532)                         & 0.2364(533)                        & 0.2779(527)                         & {\ul 0.0161(343)}    & \cellcolor[HTML]{C0C0C0}1.1518(755) & 0.0425(347)                         & \textbf{0.0148(342)} \\ 
\bottomrule[1pt]
\end{tabular}}
\vspace{-2em} 
\end{table}

\subsubsection{Hyperparameter Tuning Task}
Here we evaluate the optimal AF on 3 practical hyperparameter tuning test sets to further validate the effectiveness of our method. We utilize the surrogate benchmark implemented in JAHS-Bench-201 \cite{bansal2022jahs} to train a randomly generated neural network architecture with 2 continuous and 4 categorical hyperparameters: learning rate in $[10^{-3}, 1]$, weight decay in $[10^{-5}, 10^{-2}]$, depth multiplier in $\{1, 3, 5\}$, width multiplier in $\{4, 8, 16\}$, resolution multiplier in $\{0.25, 0.5, 1.0\}$, and training epochs in $\{1, 2, …, 200\}$. We use ReLU \cite{hahnloser2000digital} activations and stochastic gradient descent (SGD) optimizer, and do not use trivial augment \cite{muller2021trivialaugment} for data augmentation in the training pipeline. The hyperparameter tuning task is conducted on three different image classification datasets: CIFAR-10 \cite{krizhevsky2009learning}, Colorectal-Histology \cite{kather2016multi} and Fashion-MNIST \cite{xiao2017fashion}, we record the validation accuracy and total runtime as the observations of the objective and evaluation cost for each candidate configuration, respectively. For more details and implementation, please refer to \cite{bansal2022jahs}.

Table \ref{hpo exp} shows the results achieved by all AFs using different total runtimes (denoted as C, in minutes) as budgets. It can be observed that the optimal AF achieves the best results on CIFAR-10 and Fashion-MNIST datasets. In addition, we demonstrate the results within 25 and 50 total evaluations (denoted as T) when the total runtime is sufficient. It can be observed that the optimal AF can achieve the best results in most cases, which further proves that our method is still sample-efficient, while EIpu and EI-cool perform worse than EI.

\begin{table}
\centering
\caption{Means (stds) of validation accuracies obtained by different AFs on all data sets over 10 independent runs. The best mean result for each row is highlighted in bold.}\label{hpo exp}
\scalebox{0.75}{
\begin{tabular}{cccccc}
\toprule[1pt]
Data Set                              & Budget  & EI                    & EIpu                  & EI-cool             & EvolCAF                \\ \hline
\multirow{4}{*}{CIFAR-10}             & C=2000  & 0.6885(0.05)          & {0.6934(0.06)}    & 0.6849(0.05)        & \textbf{0.7495(0.04)}  \\
                                      & C=4000  & {0.7975(0.03)}    & 0.7951(0.04)          & 0.7721(0.03)        & \textbf{0.8002(0.03)}  \\
                                      & T=25    & {0.7065(0.06)}    & 0.6765(0.08)          & 0.6744(0.08)        & \textbf{0.8030(0.04)}  \\
                                      & T=50    & \textbf{0.8394(0.03)} & 0.7980(0.07)          & 0.7744(0.08)        & {0.8351(0.02)}     \\ \hline
\multirow{4}{*}{Colorectal-Histology} & C=1000  & 0.8883(0.04)          & \textbf{0.9140(0.01)} & {0.9125(0.02)}  & 0.8901(0.02)           \\
                                      & C=2000  & 0.9039(0.05)          & \textbf{0.9273(0.01)} & {0.9196(0.01)}  & 0.9158(0.01)           \\
                                      & T=25    & {0.8779(0.04)}    & 0.8592(0.08)          & 0.8588(0.09)        & \textbf{0.9072(0.02)}  \\
                                      & T=50    & {0.9040(0.02)}    & 0.8944(0.06)          & 0.8910(0.08)        & \textbf{0.9199(0.01)}  \\ \hline
\multirow{4}{*}{Fashion-MNIST}        & C=5000  & 0.9021(0.03)          & {0.9191(0.01)}    & 0.9188(0.01)        & \textbf{0.9370(0.01)}  \\
                                      & C=10000 & 0.9238(0.02)          & 0.9318(0.01)          & {0.9355(0.008)} & \textbf{0.9425(0.007)} \\
                                      & T=25    & {0.8986(0.03)}    & 0.8711(0.06)          & 0.8730(0.06)        & \textbf{0.9349(0.01)}  \\
                                      & T=50    & {0.9218(0.02)}    & 0.8810(0.06)          & 0.8928(0.05)        & \textbf{0.9445(0.002)} \\ 
\bottomrule[1pt]
\end{tabular}}
\vspace{-2em} 
\end{table}

\begin{figure}
\centering
\includegraphics[scale=0.32]{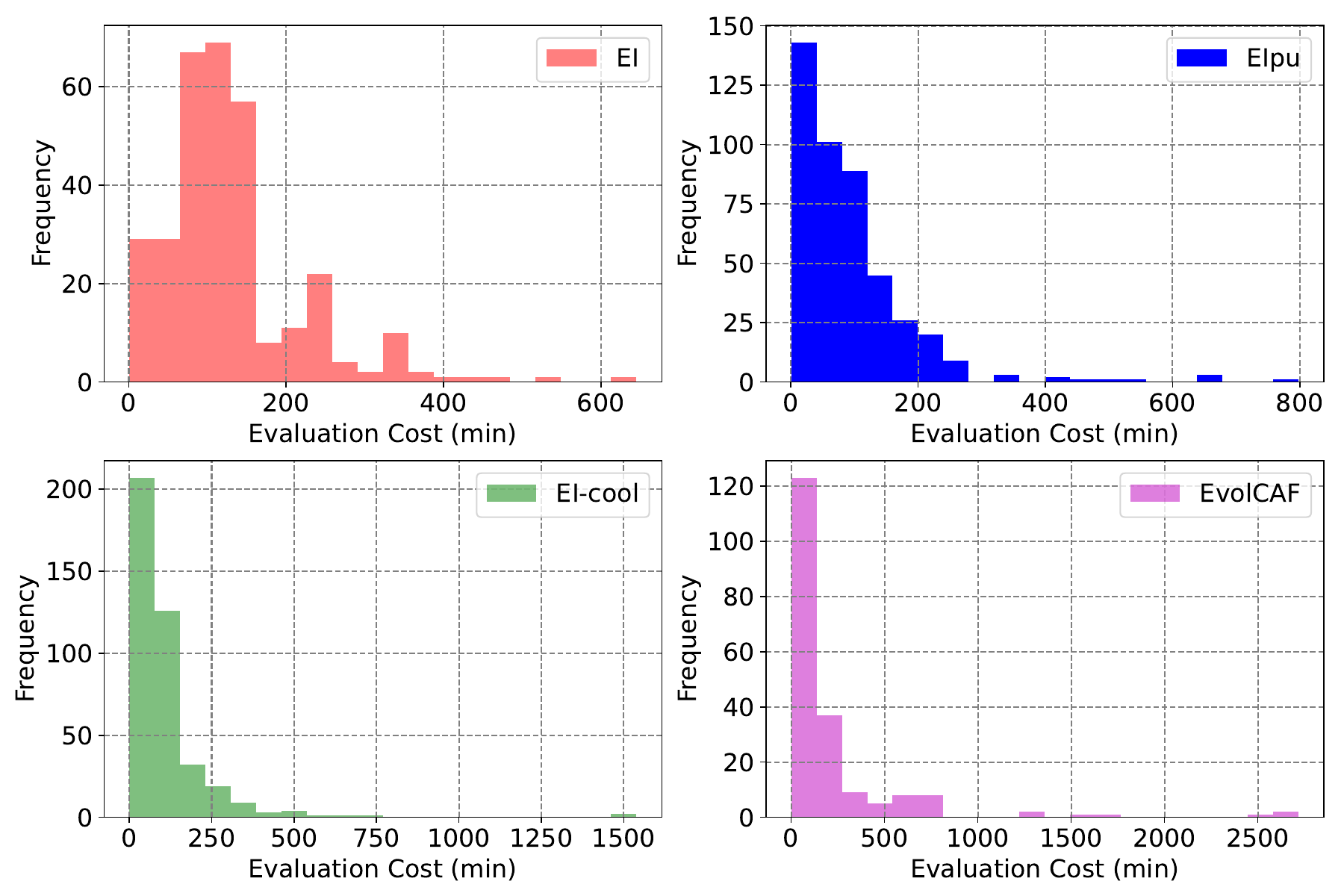}
\caption{Histograms of evaluation cost frequency collected in 10 independent runs on CIFAR-10 data set with C=4000.} \label{histogram of evaluation times}
\end{figure}

To make a further illustration, we present the histograms of evaluation cost frequency on CIFAR-10 data set with C=4000 as an example, as shown in Fig.~\ref{histogram of evaluation times}. It is evident that the majority of search frequencies of all cost-aware AFs are concentrated in cheap regions. While EIpu and EI-cool demonstrate a greater ability to explore expensive regions compared to EI, the optimal AF has the potential to explore regions that are significantly more expensive than those searched by EIpu and EI-cool, resulting in superior performance.

\section{Conclusion}
This paper introduces EvolCAF, a novel framework that integrates large language models (LLMs) with evolutionary computation (EC) to automatically design cost-aware AFs. Leveraging crossover and mutation in the algorithmic space, EvolCAF offers a novel design paradigm that significantly reduces the reliance on domain expertise and model training. The designed AF showcases novel ideas not previously explored in the existing literature on AF design, allowing for clear interpretations to provide insights into its behavior and decision-making process. Compared to the well-known EIpu and EI-cool methods designed by human experts, our approach demonstrates remarkable efficiency and generalization across various tasks, including 12 synthetic problems and 3 real-world hyperparameter tuning test sets. We have deployed our method to the latest proposed automatic heuristic design platform named EoH \cite{liu2024evolution}, the source code can be found in \url{https://github.com/FeiLiu36/EoH/tree/main/examples/user_bo_caf}.

In future work, we expect that the EvolCAF framework can be well adapted to other popular BO settings, such as high-dimensional BO, batch BO, multi-objective BO. Furthermore, we are going to explore the integration of different types of cost functions into the evolutionary process to enhance the robustness of the designed AF.

\subsubsection{Acknowledgments}
The work described in this paper was supported by the Research Grants Council of the Hong Kong Special Administrative Region, China [GRF Project No. CityU-11215723], the Natural Science Foundation of China (Project No: 62276223), and the Key Basic Research Foundation of Shenzhen, China (JCYJ20220818100005011).

\subsubsection{Disclosure of Interest}
The authors declare that they have no known competing interest that could appear to influence the work reported in this paper.

\bibliographystyle{splncs04}
\bibliography{references}

\end{document}